\documentclass[11pt]{article}

\usepackage[final]{acl}

\usepackage{times}
\usepackage{latexsym}
\usepackage{multirow}
\usepackage[table]{xcolor}
\usepackage{enumitem}
\usepackage{bbding}

\usepackage[T1]{fontenc}

\usepackage[utf8]{inputenc}
\usepackage{algorithm}
\usepackage{algorithmic}
\usepackage{amsmath}
\usepackage{booktabs} 
\usepackage{tabularx}
\usepackage{array}
\usepackage{listings}
\usepackage{xcolor}
\usepackage{graphicx}
\usepackage{microtype}

\NewDocumentCommand{\bx}
{ mO{} }{\textcolor{blue}{\textsuperscript{\textit{bx}}\textsf{\textbf{\small[#1]}}}}

\usepackage{inconsolata}

\usepackage{graphicx}

\title{BioInsight: Multi-Agent Orchestration for \\ Interactive Biomedical Knowledge Discovery}

\author{
 \textbf{Jieyi Wang\textsuperscript{1}},
 \textbf{Bingxuan Li\textsuperscript{2,3}},
 \textbf{Nanyi Jiang\textsuperscript{3}},
 \textbf{Desong Meng\textsuperscript{1}},
 \textbf{Zirui Fan\textsuperscript{3}}, \\
 \textbf{Yuxin Guo\textsuperscript{4}},
 \textbf{Jiayu Liu\textsuperscript{2}},
 \textbf{Kunlun Zhu\textsuperscript{2}},
 \textbf{Eddie Yang\textsuperscript{3}},
 \textbf{Xiusi Chen\textsuperscript{2}},
 \textbf{Pan Lu\textsuperscript{5}},
 \textbf{Bingxin Zhao\textsuperscript{3}}\thanks{Corresponding author.}
 \\
 \textsuperscript{1}Peking University,
 \textsuperscript{2}University of Illinois at Urbana-Champaign \\
 \textsuperscript{3}University of Pennsylvania
 \textsuperscript{4}Purdue University
 \textsuperscript{5}Stanford University
\\
 \small{
   {joysw@stu.pku.edu.cn, bxzhao@wharton.upenn.edu}
 }
}

\begin{document}
\maketitle
\begin{abstract}
Biomedical deep-research systems increasingly retrieve and synthesize scientific evidence, but their outputs typically collapse heterogeneous evidence into static text, making provenance difficult to inspect and reuse. We formulate evidence-centered biomedical knowledge discovery, where disease-associated protein signals are transformed into a structured evidence state connecting proteins, pathways, publications, interactions, claims, and uncertainty. We introduce \textsc{BioInsight}, a provenance-preserving multi-agent orchestration framework built around typed artifact contracts and an independent Search Agent that decouples evidence acquisition from downstream mechanistic reasoning, supporting both the citation-grounded report and an interactive evidence workspace, without independently regenerating evidence for visualization. We evaluate BioInsight on standardized biomedical QA, challenging protein-function reasoning, and end-to-end biomedical evidence synthesis. The results demonstrate that BioInsight achieves better traceability and ranking performance than standard search-augmented baselines, and suggest that biomedical AI systems should move toward provenance-preserving, interactive evidence artifacts.
\end{abstract}

\section{Introduction}
\vspace{-6pt}

Biomedical researchers increasingly use AI-generated analyses and reports to support research interpretation and decision-making~\citep{barabasi2011networkmedicine, wang2025accelerating, zhang2024leveraging}. In disease biology, this process often starts from protein-level signals, where cohort studies or experimental screens identify disease-associated proteins and researchers must decide which pathways, mechanisms, and follow-up hypotheses warrant further investigation. Such interpretation requires more than ranking proteins: protein signals must be contextualized with pathway annotations, protein–protein interaction networks, relevant literature, and drug–target evidence~\citep{menche2015diseasenetworks,yildirim2007drugtargetnetwork}. Researchers therefore need to interrogate how a conclusion was produced, which evidence supports it, where competing interpretations remain plausible, and how claims link back to source proteins and papers~\citep{gao2024empowering,gierend2024provenance}.

Existing search-augmented LLMs and deep research agents can retrieve evidence, browse the literature, and generate citation-grounded answers~\citep{jin2025searchr1,li2025webthinker,liu2025webexplorer,shao2025drtulu}. Biomedical agents such as Biomni further demonstrate that language models can coordinate tools, databases, and code execution to perform complex biomedical tasks~\citep{huang2025biomni}. However, their outputs are typically delivered as static textual reports, making it difficult for researchers to navigate the evidence underlying individual claims. To present in a form that researchers can actively inspect, interactive interfaces can make evidence, uncertainty, and provenance visible, but it's harder to ensure the accuracy of the claim. For biomedical decision-making, it requires an evidence representation that remains coherent and traceable throughout retrieval, synthesis, and user interaction~\citep{widesearch,leviathan2026generativeui}.

\begin{figure*}[!htp]
\centering
\includegraphics[width=0.88\linewidth]{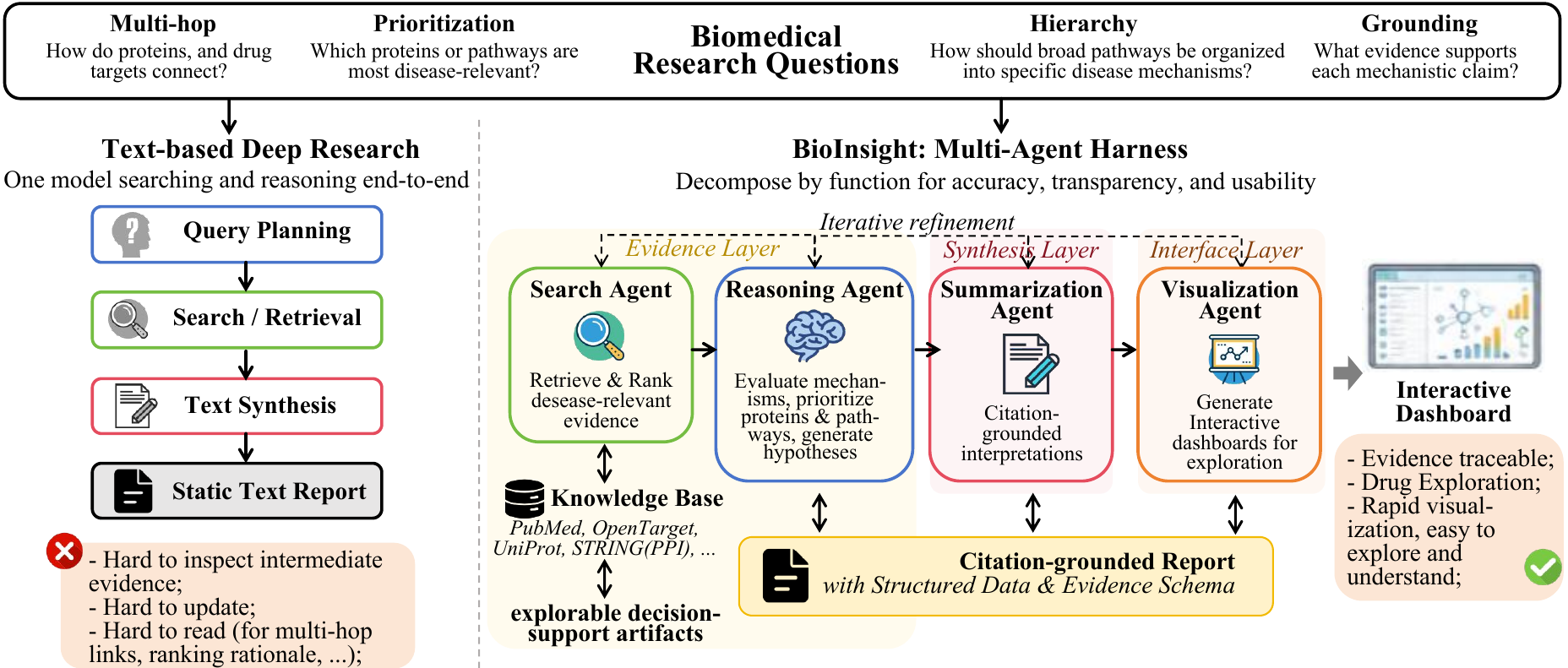}
\caption{BioInsight converts disease-centered protein evidence into an interactive evidence interface. The system uses agent-produced artifacts for search, reasoning, report writing, and dashboard construction, so users can move from high-level hypotheses to the proteins, pathways, publications, and interaction evidence behind them.}
\label{fig:RQ}
\vspace{-10pt}
\end{figure*}

We argue that a persistent evidence layer shared across retrieval, reasoning, and presentation by agent harness enginerring can help. The layer should use explicit schemas to represent proteins, pathways, publications, and their relations. It should preserve provenance by linking every downstream claim to the entities, analytical results, and source documents from which it is derived. Each reasoning, writing, and visualization agent is resuable to operate over shared evidence objects rather than reconstructing information from free-form text. For inspection, intermediate evidence can be examined independently of any particular report or interface.

We therefore formulate disease-centered protein interpretation as an evidence-centered biomedical knowledge discovery task~\citep{agrawal2018large,koscielny2017opentargets}. Given a disease name, a table of disease-associated proteins, and optional cohort metadata, the system must retrieve, organize, and synthesize heterogeneous biomedical evidence while preserving the lineage from observed protein signals to downstream mechanisms, claims, and supporting sources. Relative to search agents, deep research systems, and biomedical agents (Table~\ref{tab:concept}), the primary object is this structured evidence state rather than any single presentation format.

We propose \textbf{BioInsight}, a harness-centered multi-agent system that realizes this formulation (Fig.~\ref{fig:RQ}). An Independent Search Agent acquires, ranks, and normalizes heterogeneous biomedical evidence; typed artifact contracts maintain the resulting shared evidence state across stages; and a Reasoning Agent synthesizes mechanisms over these structured artifacts. A Writing Agent and Visualization Agent subsequently render a citation-grounded report and an interactive dashboard as two synchronized views over the same evidence state rather than independently generated endpoints. We hypothese that biomedical deep-research systems benefit from preserving evidence as structured, provenance-linked artifacts across stages rather than free-form generation. We test this through standardized biomedical QA, protein-function reasoning, and end-to-end evaluation of generated reports and dashboards. Across these settings, BioInsight achieves the best or tied-best QA performance, the highest expert score on BioInsight-100, and stronger expert ratings for traceability, ranking quality, and dashboard usability in disease-level interpretation. To summarize, our contributions are threefold:
\begin{itemize}[noitemsep, topsep=0pt]
    \item We formulate disease-centered protein interpretation as an evidence-centered biomedical knowledge discovery task in which heterogeneous evidence is organized into a provenance-preserving state that supports continued synthesis and inspection.

    \item We propose BioInsight, a harness-centered multi-agent architecture that uses typed artifact contracts and an Independent Search Agent to decouple evidence acquisition from mechanistic reasoning while maintaining provenance across retrieval, reasoning, reporting, and visualization.

    \item We evaluate whether the proposed orchestration preserves biomedical capability while improving evidence synthesis and provenance. Our evaluation combines standardized biomedical QA, the BioInsight-1k protein-function reasoning benchmark, controlled same-base-model and ablation comparisons, a dedicated faithfulness analysis of evidence preservation, and expert assessment of end-to-end disease interpretation.
\end{itemize}

\section{Related Work}
\vspace{-6pt}

\subsection{Search and Deep Research Agents}
\vspace{-4pt}
Recent language agents support multi-step search and long-form research writing~\citep{yao2023react, wu2024autogen, li-etal-2025-metal, liu2026osexpert, li2026pearl, Li_2026_CVPR}. Search-R1 and ASearcher train models to interleave reasoning with search~\citep{jin2025searchr1,gao2025asearcher}, while WebThinker, WebExplorer, Tongyi DeepResearch, and DR-Tulu focus on well-attributed answers or reports over retrieved evidence~\citep{li2025webthinker,liu2025webexplorer,tongyi2025deepresearch,shao2025drtulu}. Evaluation work on RAG and self-reflective retrieval further emphasizes grounding and evidence use, not only final text quality~\citep{es2024ragas,asai2024selfrag,yan2024crag, zhu-etal-2025-safescientist,yu2025tinyscientist}. These systems typically couple retrieval and reasoning through free-form context: search is an action within an end-to-end generation loop, and intermediate evidence is not retained as a typed, reusable state across stages. BioInsight instead treats evidence acquisition as an independent stage whose outputs are normalized into inspectable artifacts before downstream reasoning.

\begin{table}[t]
\tiny
\centering
\small
\setlength{\tabcolsep}{1pt}
\begin{tabular}{lcccc}
\toprule
\textbf{Design Aspect} & \textbf{Search} & \textbf{Deep Res.} & \textbf{Bio. Agent} & \textbf{Ours} \\
\midrule
Evidence retrieval & \Checkmark & \Checkmark & \Checkmark & \Checkmark \\
Tool integration & Partial & Partial & \Checkmark & \Checkmark \\
Typed artifacts & $\times$ & $\times$ & Partial & \Checkmark \\
Cross-stage provenance & $\times$ & Partial & Partial & \Checkmark \\
Interface state & $\times$ & $\times$ & $\times$ & \Checkmark \\
\bottomrule
\end{tabular}
\caption{Architecture comparison of BioInsight with search agents (e.g., Search-R1\cite{jin2025searchr1}), deep research systems (e.g., DR-Tulu\cite{shao2025drtulu}, WebThinker\cite{li2025webthinker}), and biomedical agents (e.g., Biomni\cite{huang2025biomni}).}
\label{fig:concept}
\vspace{-8pt}
\end{table}

\subsection{Biomedical Agents for Evidence Synthesis}
\vspace{-4pt}
Biomedical LLM agents connect language models with literature, databases, tools, and code execution environments~\cite{sokolova2025evidence}. Biomni demonstrates broad biomedical task automation through tool use and planning~\citep{huang2025biomni}; biomedical RAG and search systems improve access to PubMed and domain knowledge bases before generation~\citep{bi2025bioragent,liu2025biomedsearch}. These systems provide strong tool and database integration, but retrieved evidence is still often treated as transient context for question answering or task completion. BioInsight targets a narrower setting---disease-centered protein interpretation---where protein association signals must be connected to pathways, publications, PPI modules, drug--target context, and uncertainty~\citep{koscielny2017opentargets,agrawal2018large}. As summarized in Table~\ref{fig:concept}, the distinguishing design is not broader tool coverage alone, but an explicit evidence layer that keeps these heterogeneous sources typed, provenance-linked, and reusable across reasoning and presentation.

\subsection{Interactive Scientific Interfaces}
\vspace{-4pt}
Visual analytics and generative interface systems explore how scientific information can be inspected beyond static text~\citep{sosa2020literaturebased}. PaperVoyager converts papers into executable interactive systems~\citep{dai2026papervoyager}, and recent generative UI work shows that LLMs can produce task-specific interfaces~\citep{leviathan2026generativeui,google2026a2ui}. Biomedical settings impose a stricter constraint: visual elements must remain tied to heterogeneous evidence such as proteins, pathways, publications, interactions, and drug context~\citep{ehlers2025biologicalnetworkvis}. BioInsight therefore does not treat the interface as an independent generative endpoint. The dashboard is a synchronized view of the same typed evidence state used to write the report, so interactivity inherits provenance from the orchestration layer rather than compensating for missing structure after generation.

\section{Method}
\vspace{-8pt}
We present \textbf{BioInsight}, a harness-centered multi-agent system for \textbf{wide biomedical evidence synthesis} in disease-centered protein interpretation. BioInsight decomposes disease-centered protein evidence into an interactive decision-support workspace. The report is one artifact in this process, not the endpoint. The system first builds structured evidence objects from protein associations, pathway enrichment, literature retrieval, and interaction data; it then uses those objects to generate both a citation-grounded narrative and a dashboard schema. The rendered dashboard exposes the same evidence chain used by the report, allowing users to inspect claims, proteins, citations, and uncertainty from multiple views.

\subsection{Task Formulation}

We define its task as evidence-centered interactive interface generation: given disease-associated protein signals, multiple agents iteratively retrieve, rank, reason over, write about, and visualize biomedical evidence to produce explorable decision-support artifacts. The harness coordinates these agents through typed artifact contracts, so each refinement step can reuse, revise, or expose upstream evidence without losing provenance. An input instance is
\[
x = (d, P, M, K),
\]
where \(d\) is a disease name, \(P\) is a disease-associated protein table, \(M\) is optional cohort metadata, and \(K\) denotes external biomedical knowledge resources. The protein table contains protein symbols and statistical association fields such as hazard ratios, confidence intervals, and \(P\) values.

The system produces
\[
Y = (T, E, N, R, S, H),
\]
where \(T\) is a ranked pathway table, \(E\) is a set of evidence packets, \(N\) contains pathway- and protein-level reasoning notes, \(R\) is a citation-grounded report, \(S\) is a dashboard schema, and \(H\) is the rendered interactive dashboard. Search and planning agents construct \(T\) and \(E\); reasoning and writing agents refine them into \(N\) and \(R\); the visualization agent converts the same evidence base into \(S\) and \(H\). \(H\) is the primary user-facing artifact, while \(T\), \(E\), \(N\), \(R\), and \(S\) provide its provenance. Each downstream artifact keeps references to the upstream proteins, pathways, statistics, evidence packets, and citations from which it was derived.

\subsection{Multi-Agent Harness Design}
\vspace{-4pt}
\subsubsection{Overview}

BioInsight has three layers. The evidence layer performs pathway planning and publication retrieval. The synthesis layer produces structured reasoning notes and a citation-grounded report. The interface layer converts the same evidence objects into an interactive dashboard. A central harness schedules these stages, validates required fields, and passes only typed artifacts between agents. Given a disease name, protein table, and cohort metadata, the system sequentially plans pathway-level hypotheses, retrieves biomedical evidence, synthesizes structured mechanisms, formats citations, and renders the final report and interactive dashboard.  All LLM agents use GPT-4o as the base model.

This organization keeps generated interactive interface tied to evidence. The Visualization Agent does not invent new nodes or claims. It receives structured outputs from earlier stages and renders them as an evidence workspace. If a pathway, protein, or publication is missing from the artifacts, it cannot appear as a supported item in the dashboard.

\subsubsection{Evidence Planning and Retrieval}

\paragraph{Pathway Planning}

The Planning Agent identifies candidate biological mechanisms from the input protein set. It maps disease-associated proteins to enriched biological terms using g:Profiler. Near-duplicate pathway names are removed with BioBERT embedding similarity after generic pathway phrases are stripped. The remaining pathways are ranked by combining enrichment strength with disease-specific literature support.

For pathway \(t\), let \(p(t)\) denote its enrichment \(P\) value and \(L(t)\) denote the literature relevance score returned by the Search Agent. We normalize these values across candidate pathways to obtain \(P_{\mathrm{norm}}(t)\) and \(L_{\mathrm{norm}}(t)\), where lower \(P_{\mathrm{norm}}(t)\) indicates stronger enrichment and higher \(L_{\mathrm{norm}}(t)\) indicates stronger disease-specific literature support. The pathway score is
\[
S_{\mathrm{path}}(t)
=
0.4 \cdot P_{\mathrm{norm}}(t)
+ 0.6 \cdot (1 - L_{\mathrm{norm}}(t)).
\]

Lower scores are prioritized. This favors pathways that are both statistically supported by the input protein set and grounded in disease-relevant literature.

\paragraph{Publication Retrieval and Scoring}

For each candidate pathway, the Search Agent builds disease-pathway queries from the disease name and pathway name. It retrieves publications from PubMed and Semantic Scholar, then normalizes them into evidence packets. Each publication is scored using lexical relevance, semantic relevance, citation impact, and journal weight. For query \(q=(d,t)\) and article \(a\), the raw score is
\[
\begin{aligned}
S_{\mathrm{raw}}(q,a)
=&\ 0.45K(q,a) + 0.35E(q,a) \\
&+ 0.20C(a),
\end{aligned}
\]
where \(K(q,a)\) measures disease-pathway keyword matches, \(E(q,a)\) is BioBERT semantic similarity between the query and title-abstract text, and \(C(a)\) is a log-scaled citation-count score. The final score applies a bounded journal weight:
\[
S_{\mathrm{pub}}(q,a)
=
\operatorname{clip}
\left(
S_{\mathrm{raw}}(q,a) \cdot J(a), 0, 2.2
\right),
\]
where \(J(a)\) is derived from the publication venue. Publications with \(S_{\mathrm{pub}} \geq 0.25\) are retained as validated evidence. Each retained item stores the query, metadata, relevance score, PMID when available, and the pathway for which it was retrieved.

\subsubsection{Evidence Synthesis}

For each selected pathway, the Reasoning Agent receives the disease name, pathway identifier, pathway description, intersecting proteins, original association statistics, validated publication packets, protein function records, and PPI records when available. It returns structured reasoning notes instead of final prose. Each note contains a pathway interpretation, disease relevance, key protein summaries, PPI-module explanations, uncertainty statements, and citation links. The full reasoning-note schema is provided in Appendix~A.

Protein--protein interactions are represented as graph evidence. For proteins intersecting a pathway, BioInsight builds an undirected weighted graph from retrieved PPI edges, identifies connected components, and ranks components by internal edge weight and size. The top components are used for cluster-level reasoning; disconnected proteins are analyzed individually using protein function annotations. This separates isolated protein evidence from coherent interaction modules within a disease-relevant pathway.

\begin{figure*}[t]
\centering
\includegraphics[width=0.95\linewidth]{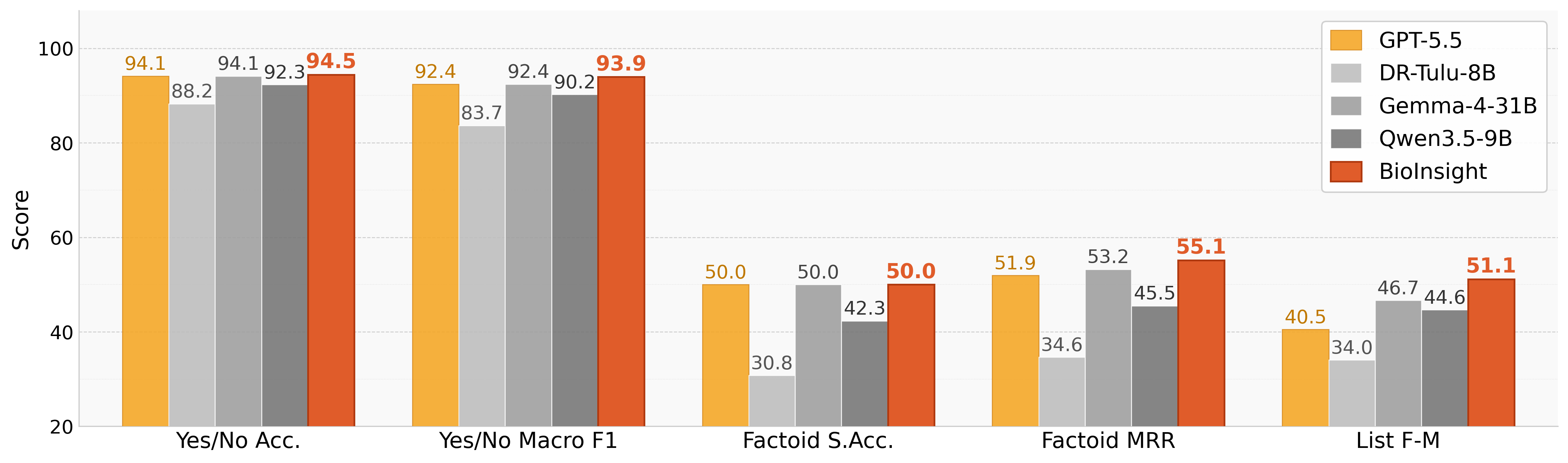}
\caption{Results on BioASQ Phase B Exact Answer task, Batch 1. All five metrics are higher-is-better. BioInsight achieves the best performance across all five exact-answer metrics.}
\label{fig:BioASQ}
\end{figure*}

\subsubsection{Evidence-Centered Interactive Interface Generation}

The Writing Agent organizes reasoning notes into a scientific report. The report includes a disease introduction, cohort and protein summary, ranked pathway table, pathway-level analyses, protein-level explanations, PPI summaries and links, optional drug or target context, and citation identifiers. Then, the Visualization Agent converts the organized insights, original protein table, and cached evidence artifacts into a dashboard schema. Pathways, proteins, publications, PPI edges, and optional drug-related edges are extracted according to predefined schemas. Optional language-model summarization is used only to shorten long display text; it does not create new evidence objects.

The dashboard contains a pathway ranking panel, a graph linking pathways, proteins, publications, PPI edges, and optional drug-related evidence, and detail panels for selected nodes or edges. Users can inspect enriched proteins for a pathway, view original association statistics for a protein, open PMID-linked publication evidence, and filter the graph by edge type. The output is therefore an explorable decision-support artifact built from the same evidence that supports the generated report.

\section{Experiments and Results}
We evaluate BioInsight along the properties introduced by artifact-centered orchestration. We ask four research questions:
\begin{itemize}[noitemsep, topsep=2pt]
    \item \textbf{RQ1 (Grounding).} Does BioInsight preserve the biomedical retrieval and entity-selection capabilities required for downstream evidence synthesis?
    \item \textbf{RQ2 (Cross-source reasoning).} Can BioInsight integrate heterogeneous evidence for challenging protein-centered reasoning?
    \item \textbf{RQ3 (Orchestration contribution).} Does artifact-centered orchestration improve evidence synthesis independently of the underlying foundation model?
    \item \textbf{RQ4 (Provenance / faithfulness).} Does the typed-artifact contract preserve evidence across report and interface generation?
\end{itemize}

\subsection{Baselines and Setup}

We use task-specific controls. For RQ1 and RQ2, we compare BioInsight with GPT-5.5, DR-Tulu-8B, Gemma-4-31B, and Qwen3.5-9B, covering general-purpose, biomedical/domain-aligned, and open-weight systems. For RQ3, the primary comparison is same-base-model: GPT-4o+Search vs.\ BioInsight~(GPT-4o), GPT-5.5 vs.\ BioInsight~(GPT-5.5), and BioInsight without the Independent Search Agent (w/o ISA). We additionally report external comparisons against strong search-enabled systems (GPT-5.5+Search, Gemini Deep Research) and open-weight baselines on five disease cases (Appendix~\ref{app:diseases}). For RQ4, we compare BioInsight dashboards with report-first GPT-5.5 pipelines under matched evidence access, including a typed-input-package control that receives the same retrieved evidence without BioInsight orchestration. All systems use the same disease names, protein association tables, and task instructions where applicable. Additional model settings and retrieval budgets are in Appendix~\ref{app:experimental-setup}.

\subsection{Grounding Accuracy in Standardized Biomedical QA}
\vspace{-4pt}
Before a multi-agent workflow can maintain a reusable evidence state, it must preserve exact biomedical answering, entity normalization, and evidence-supported item selection. We therefore treat BioASQ Phase B (yes/no, factoid, and list) as a \textbf{prerequisite grounding check}.

Figure~\ref{fig:BioASQ} shows that BioInsight maintains competitive exact-answer performance without sacrificing QA ability. Confidence intervals are wide, so point-estimate gaps should be interpreted cautiously. The most consistent signal is on List F-measure (51.10 vs.\ 40.49 for GPT-5.5), which aligns with the Independent Search Agent's role in multi-entity retrieval and evidence aggregation. 

\begin{figure}[ht]
\centering
\includegraphics[width=0.85\linewidth]{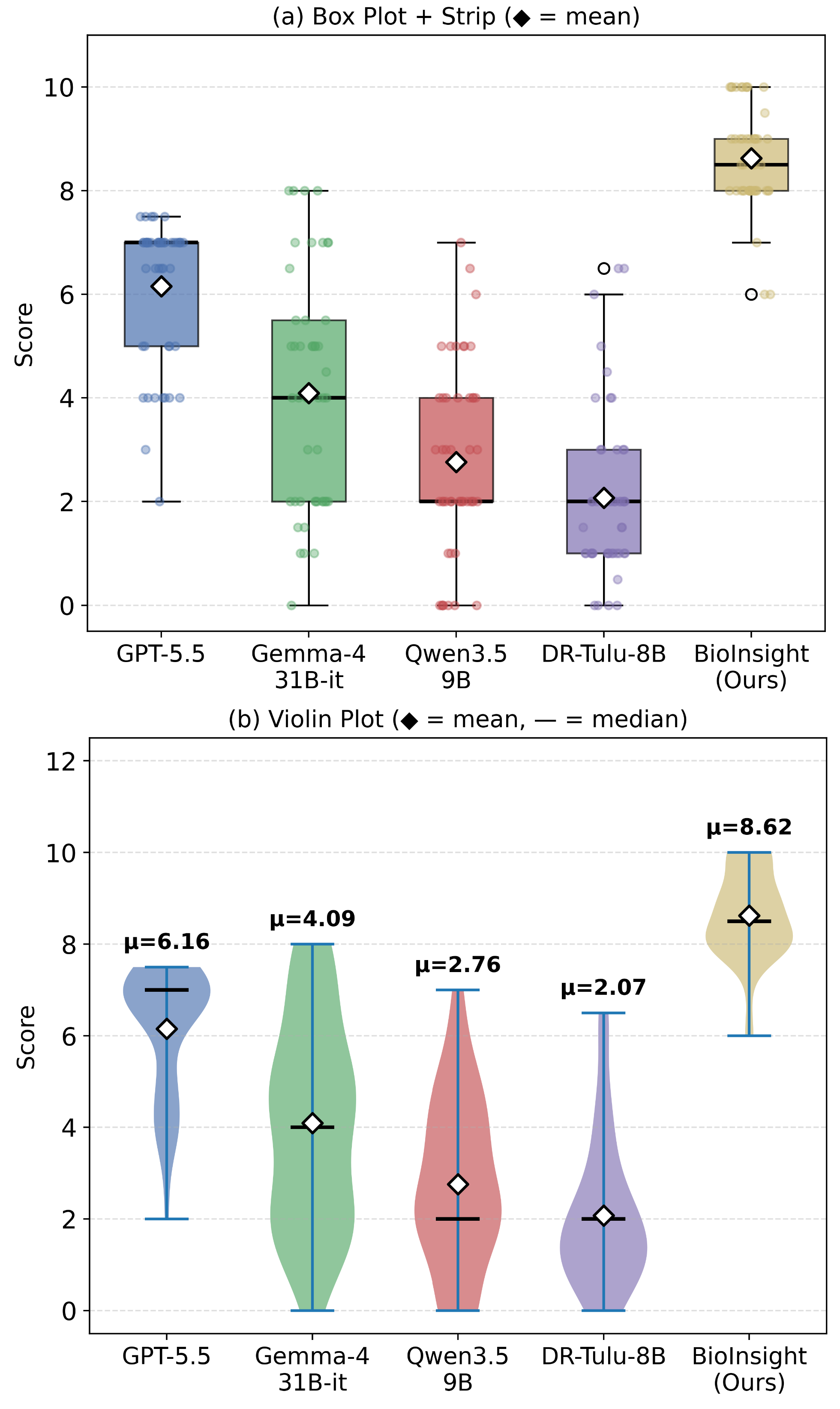}
\caption{Evaluation score distributions on the BioInsight-100 benchmark, a challenging subset of protein-function analysis questions from BioInsight-1k. Box plots with individual data points (a) and violin plots (b) show the distribution of 0--10 scores.}
\label{fig:pfq}
\vspace{-10pt}
\end{figure}

\begin{figure}[htp]
\centering
\includegraphics[width=1\linewidth]{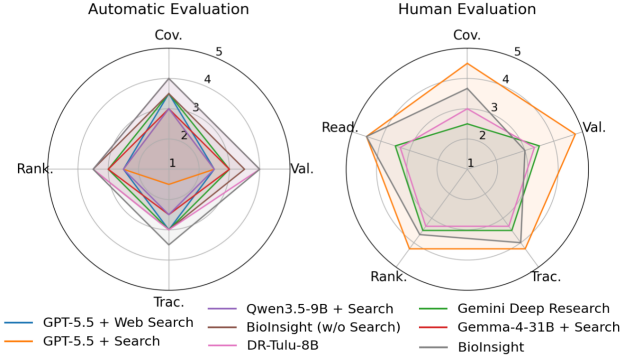}
\caption{Automatic and human evaluation of end-to-end biomedical evidence synthesis reports on Atrial fibrillation and flutter. Cov., Val., Trac., Rank. and Read. respectively means coverage, biomedical validity, evidence traceability, ranking and research depth, and user usability. Higher scores indicate better performance.}
\label{fig:report_eval}
\vspace{-10pt}
\end{figure}

\subsection{Cross-Source Protein Reasoning}
\vspace{-4pt}
Disease-centered protein interpretation further requires integrating functional annotations, PPIs, pathway context, disease mechanisms, and ranked explanations across heterogeneous sources. We construct BioInsight-1k from UniProt and STRING evidence, use GPT-5.5 to propose candidate questions, and have biomedical experts select 100 challenging cross-evidence items to form BioInsight-100. The benchmark is not a factual lookup test: questions require cross-source reasoning and prioritization, as verified by two biomedical experts. All systems access the same underlying resources under the same protocol.

Figure~\ref{fig:pfq} reports expert scores. BioInsight obtains the highest mean score of 8.62, with answers concentrated in the high-score range, indicating more stable protein-centered reasoning and evidence ranking than general-purpose and open-weight baselines. 

\subsection{Orchestration Contribution}
To evaluate whether typed artifacts and the Independent Search Agent improve evidence synthesis \textbf{independently of the foundation model}, we evaluate end-to-end disease-centered synthesis on coverage, biomedical validity, evidence traceability, and ranking/research depth~\citep{page2021prisma, clark2024generative, flemyng2025position}(rubric details in Appendix~\ref{app:human-evaluation}).

\begin{table*}[t]
\footnotesize
\centering
\caption{Automatic and human evaluation of end-to-end biomedical evidence synthesis reports (except AF).}
\begin{tabular}{ccccccccccc}
\hline
Disease & Models                  & \multicolumn{4}{c}{Automatic Evaluation}      & \multicolumn{5}{c}{Human Evaluation} \\ \cline{3-11} 
        &                         & Cov.      & Val.      & Trac.     & Rank.     & Cov.  & Val. & Trac. & Rank. & Read. \\ \hline
AD      & GPT5.5 + Web            & 4.0       & 4.0       & 3.0       & 3.5       & 4.00  & 4.75 & 4.50  & 4.25  & 4.50  \\
        & Gemini Deep Search      & 4.0       & 3.0       & 3.0       & 3.5       & 4.50  & 3.50 & 4.00  & 4.00  & 4.50  \\
        & DR-Tulu-8B              & 3.0       & 2.5       & 2.0       & 2.5       & 3.00  & 4.00 & 3.67  & 3.33  & 3.67  \\
        & Qwen3.5-9B              & 3.0       & 2.5       & 2.0       & 2.5       & -     & -    & -     & -     & -     \\
        & BioInsight (w/o Search) & 4.0 (3.5) & 4.0 (3.0) & 3.5 (3.0) & 4.0 (3.0) & 3.75  & 4.00 & 4.50  & 4.00  & 4.50  \\ \hline
CKD     & GPT5.5 + Web            & 3.5       & 4.0       & 3.5       & 3.5       & 4.33  & 5.00 & 4.33  & 3.67  & 4.00  \\
        & Gemini Deep Search      & 4.0       & 4.0       & 2.5       & 3.5       & 5.00  & 4.50 & 3.50  & 4.00  & 4.50  \\
        & DR-Tulu-8B              & 3.0       & 2.5       & 1.5       & 2.5       & 2.33  & 3.00 & 2.00  & 2.33  & 2.33  \\
        & Qwen3.5-9B              & 3.0       & 3.0       & 2.5       & 3.0       &       &      &       &       &       \\
        & BioInsight (w/o Search) & 3.5 (3.5) & 4.0 (3.0) & 3.5 (3.0) & 3.5 (3.0) & 4.67  & 4.00 & 3.67  & 4.00  & 4.50  \\ \hline
MDD     & GPT5.5 + Web            & 3.5       & 4.0       & 3.5       & 4.0       & 4.33  & 4.67 & 4.33  & 3.67  & 4.33  \\
        & Gemini Deep Search      & 4.0       & 3.0       & 2.5       & 3.5       & 4.50  & 4.50 & 3.50  & 4.50  & 4.50  \\
        & DR-Tulu-8B              & 2.5       & 2.5       & 1.5       & 2.0       & 3.00  & 2.33 & 3.33  & 2.33  & 2.00  \\
        & Qwen3.5-9B              & 3.0       & 3.0       & 2.5       & 3.0       & -     & -    & -     & -     & -     \\
        & BioInsight (w/o Search) & 4.0 (3.5) & 4.0 (3.0) & 3.5 (2.5) & 4.0 (3.0) & 3.67  & 3.00 & 3.00  & 3.33  & 4.50  \\ \hline
RA      & GPT5.5 + Web            & 4.0       & 4.0       & 3.0       & 3.5       & 4.67  & 4.67 & 4.33  & 3.67  & 4.67  \\
        & Gemini Deep Search      & 3.5       & 3.0       & 2.5       & 3.0       & 4.50  & 4.50 & 3.50  & 3.50  & 4.50  \\
        & DR-Tulu-8B              & 3.0       & 2.0       & 1.5       & 2.5       & 2.33  & 3.33 & 2.33  & 3.00  & 2.33  \\
        & Qwen3.5-9B              & 3.0       & 2.5       & 2.0       & 2.5       & -     & -    & -     & -     &       \\
        & BioInsight (w/o Search) & 4.0 (3.5) & 4.5 (3.0) & 3.5 (3.0) & 4.0 (3.0) & 3.33  & 3.50 & 3.33  & 3.67  & 4.83  \\ \hline

\end{tabular}
\label{tab:reports}
\end{table*}

For automatic evaluation, we score all six systems on four dimensions twice: coverage, biomedical validity, evidence grounding and traceability, and prioritization and research depth. To control the result not influenced by searching ability, all "+ Search" means that the generation is based on the literature searched by BioInsight. For human evaluation, we select the four strongest or most representative systems from the automatic study: GPT-5.5 + Search, Gemini Deep Research, DR-Tulu-8B, and BioInsight. Experts assess five dimensions, with an addition of readability. The evaluation results of Atrial fibrillation and flutter are shown in Figure \ref{fig:report_eval}, and that of other diseases are listed in Table \ref{tab:reports}. 

As an external check, Figure~\ref{fig:report_eval}\footnote{An anonymized demo is available at \url{http://3.148.244.109:5000/}.} and Table~\ref{tab:reports} compare BioInsight with strong search-enabled and open-weight systems across five diseases (Appendix~\ref{app:diseases}), covering different biomedical decision regimes, including Alzheimer's disease (AD), depression (MDD), atrial fibrillation and flutter (AF), chronic kidney disease (CKD), and rheumatoid arthritis (RA).. BioInsight w/o ISA sits between the search-only baseline and full BioInsight, showing that freezing and normalizing evidence before reasoning contributes to the gain---not merely using a stronger model.
BioInsight remains competitive on coverage and validity and strongest on automatic traceability and ranking; the separate w/o ISA rows show consistent drops when evidence acquisition is not decoupled. Biomedical validity is less separated among frontier systems, consistent with validity depending more on the base model, while traceability depends more on workflow structure. Expert ratings likewise favor BioInsight on traceable synthesis, ranking, and usability. These external results support practical utility.

\subsection{Provenance and Faithfulness}
\vspace{-4pt}
To test whether typed artifact contracts preserve evidence across report and interface generation. For each of five diseases, we construct a gold reference of grounded proteins, pathway overlaps, PMIDs, STRING PPIs, and DGIdb drug links. We evaluate unsupported entities (M1\(\downarrow\)), unsupported relations (M2\(\downarrow\)), pathway omission (M3\(\downarrow\)), and citation coverage (M4\(\uparrow\)).

\begin{table*}[t]
\centering
\caption{Faithfulness evaluation with 95\% confidence intervals. Once strong models receive clean typed artifacts, unsupported entities/relations are already rare; BioInsight's main advantage is reducing evidence loss (pathway omission, citation coverage) across downstream generation.}
\small
\setlength{\tabcolsep}{3pt}
\begin{tabular}{lcccc}
\toprule
System & M1 Ent.$\downarrow$ & M2 Rel.$\downarrow$ & M3 Path.$\downarrow$ & M4 Cite.$\uparrow$ \\
\midrule
BioInsight (dashboard) & 0.000\,[0.00,0.00] & 0.000\,[0.00,0.00] & 0.110\,[0.05,0.14] & 0.334\,[0.17,0.56] \\
GPT-5.5 (typed package) report & 0.001\,[0.00,0.00] & 0.002\,[0.00,0.01] & 0.268\,[0.16,0.30] & 0.245\,[0.15,0.39] \\
GPT-5.5 (raw) report & 0.345\,[0.22,0.49] & 0.842\,[0.75,0.93] & 0.879\,[0.85,0.91] & 0.088\,[0.02,0.18] \\
GPT-5.5 (raw) dashboard & 0.399\,[0.29,0.52] & 0.458\,[0.20,0.70] & 0.947\,[0.92,0.98] & 0.039\,[0.01,0.07] \\
\bottomrule
\end{tabular}
\label{tab:faithfulness}
\end{table*}

Table~\ref{tab:faithfulness} yields a precise conclusion. Raw report/dashboard pipelines fabricate many unsupported entities and relations and lose most pathway and citation evidence. Giving GPT-5.5 the same typed evidence package already nearly eliminates unsupported entities and relations (M1/M2\(\approx 0\)), showing that hallucination reduction alone is not BioInsight's distinctive contribution. Relative to this strong typed-input control, BioInsight still substantially lowers pathway omission (0.11 vs.\ 0.27) and raises citation coverage (0.33 vs.\ 0.25), while keeping unsupported entities and relations at zero. Thus, \textbf{typed artifacts primarily improve evidence preservation across stages, not merely hallucination reduction}---exactly the property predicted by the shared evidence state and containment invariant.

\begin{figure*}[ht!]
\centering
\includegraphics[width=0.95\textwidth]{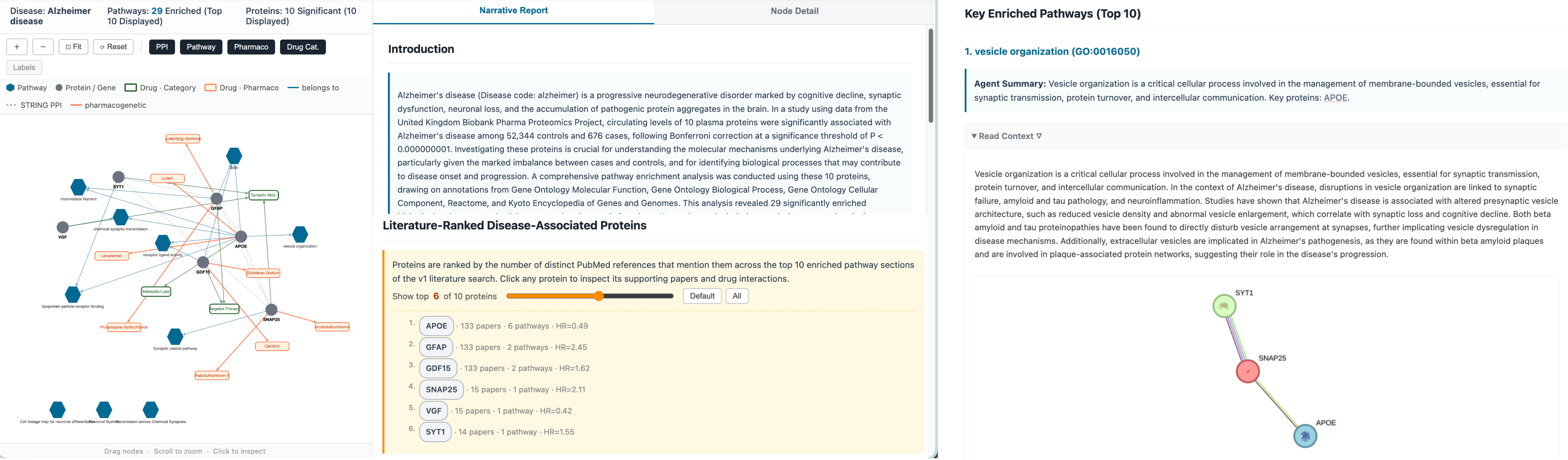}
\caption{Case study on Alzheimer's Disease.}
\label{fig:Casestudy}
\vspace{-10pt}
\end{figure*}

Furthermore, BioInsight receives the strongest expert ratings for traceable synthesis, ranking, and usability. Expert comments further indicate that BioInsight is the only system that consistently surfaces disease-association evidence, which gives researchers concrete hints for disease--protein analysis rather than only a general biological summary. The interactive interface is also recognized as useful for checking pathways, proteins, and citations. At the same time, experts are willing to use clear and well-organized Markdown reports, suggesting that the value of the interactive interface comes from evidence traceability and navigation rather than visual presentation alone. Besides, as information passes through retrieval, reasoning, writing, and visualization stages, some coverage can be lost or compressed. In our results, this did not prevent BioInsight from producing useful reports, but it points to an important open question for future deep research: more retrieved information is not always better if it cannot be ranked, grounded, and presented in a form that researchers can inspect.

Overall, the three evaluations show complementary strengths. BioInsight improves exact biomedical answering on BioASQ, achieves stronger cross-evidence protein-function reasoning on BioInsight-100, and produces more traceable and usable end-to-end biomedical evidence synthesis reports. These results support the central claim that harness-centered artifact contracts can improve biomedical evidence synthesis beyond single-step search or fluent long-form generation.

\vspace{-6pt}
\section{Case Study}
\vspace{-5pt}
We use Alzheimer's disease (AD) to illustrate BioInsight as an evidence-centered interface, not a single-step report generator. AD is a suitable case because protein signals must be interpreted across synaptic dysfunction, lipid metabolism, glial activation, axonal injury, and neurodegeneration instead of one dominant pathway.

Given the disease name, cohort metadata, and 10 significant plasma proteins, BioInsight builds an artifact chain from protein statistics to pathways, publications, reasoning notes, and the dashboard in Figure~\ref{fig:Casestudy}. The Planning Agent identifies enriched candidate pathways, the Search Agent retrieves disease-specific literature, the Reasoning Agent integrates pathway, protein, PPI, and drug-related evidence, and the Visualization Agent exposes the resulting evidence structure for inspection.

The resulting workspace keeps the interpretation grounded in the observed protein signals. APOE appears as a cross-pathway driver linking lipid receptor biology, vesicle organization, synaptic processes, and axon-related hypotheses. GFAP and NEFL support glial and axonal-injury interpretations, while SNAP25 and SYT1 form a presynaptic vesicle and chemical synapse module. Proteins with weaker pathway or literature support remain visible but are treated as exploratory.

This case highlights the role of the interactive artifact. A pathway can be statistically enriched but weakly supported by AD-specific literature, while another may have moderate enrichment but clearer disease relevance. By combining enrichment evidence with publication support and making both visible, BioInsight turns a static disease report into an auditable path from proteins to pathways, citations, mechanisms, and follow-up hypotheses.

\vspace{-6pt}

\section{Conclusion}
\vspace{-6pt}
We introduced \textsc{BioInsight}, a harness-centered multi-agent system for disease-centered protein interpretation. By enforcing artifact contracts between retrieval, reasoning, writing, and dashboard construction, BioInsight exposes protein, pathway, publication, and citation links that are usually hidden in end-to-end biomedical agents. Across BioASQ, BioInsight-100, and report-level expert evaluation, the system improves exact answering, protein-function reasoning, and traceable report synthesis. These results suggest that more biomedical evidence synthesis benefits from structured, auditable intermediate artifacts rather than fluent generation alone.

\section*{Limitations}
BioInsight is designed to support biomedical research interpretation and hypothesis generation, not clinical diagnosis, treatment selection, or other forms of clinical decision-making. 
A primary ethical risk is that users may overinterpret automatically generated pathway, protein, or drug--target explanations as validated biological mechanisms or therapeutic conclusions. 
Although BioInsight grounds its outputs in retrieved publications and exposes intermediate evidence through structured artifacts and dashboards, the underlying evidence may still be incomplete or noisy. 
Retrieval can miss relevant studies, select papers that are topically related but mechanistically weak, or suffer from protein synonym ambiguity, incomplete database coverage, and noisy input protein associations. 
As a result, BioInsight may produce incomplete evidence summaries, uncertain mechanistic links, or hypotheses that require further validation.

To mitigate these risks, BioInsight preserves citation links, protein-level statistics, intermediate artifacts, uncertainty notes, and failure-handling behaviors, allowing users to inspect how each claim is supported. 
However, these mechanisms are intended to support expert review rather than replace it. 
All outputs should be reviewed by domain experts and validated through independent biomedical analysis before being used to guide downstream experimental, translational, or clinical decisions.



\bibliography{custom}

\appendix

\section{Implementation and Artifact Details}
\label{app:implementation}

This appendix expands the implementation details behind the BioInsight harness. It is organized around the same artifact chain used in the method section: external resources, typed intermediate artifacts, implementation parameters, and failure handling.

\subsection{External Biomedical Knowledge Resources}
BioInsight consumes external biomedical knowledge through resource-specific modules. Each resource has a defined role, which prevents heterogeneous evidence from being merged into an opaque retrieval result.

\begin{table*}[htbp]
\centering
\small
\caption{External biomedical knowledge resources and their roles in the harness.}
\label{tab:resources}
\begin{tabularx}{\textwidth}{
  >{\raggedright\arraybackslash}p{0.18\textwidth}
  >{\raggedright\arraybackslash}X
  >{\raggedright\arraybackslash}X
}
\toprule
\textbf{Resource} & \textbf{Role in harness} & \textbf{Used by} \\
\midrule
g:Profiler & Pathway enrichment over the input protein set & Planning Agent \\
PubMed & Disease-pathway evidence retrieval and reference normalization & Planning Agent, Reasoning Agent, Citation Formatter \\
Semantic Scholar & Literature ranking, citation-count support, and optional snippet-level evidence retrieval & Planning Agent, Query Agent \\
STRING & Protein-protein interaction networks, figures, and external links & Reasoning Agent, Visualization Agent \\
UniProt-derived knowledge base & Protein function descriptions & Query Agent, Reasoning Agent \\
Open Targets & Target descriptions, drug context, bibliography, and associated diseases & Query Agent, Visualization Agent \\
DGIdb & Pharmacogenetic drug-protein edges & Visualization Agent \\
\bottomrule
\end{tabularx}
\end{table*}

\subsection{Intermediate Artifacts and Reasoning Notes}

BioInsight stores pathway rankings, evidence packets, reasoning notes, citation-linked drafts, network views, dashboard schemas, and rendered dashboards as separate artifacts. These artifacts make the system inspectable at several points: researchers can examine enriched pathways before reading the final narrative, trace report claims back to proteins and publications, and check whether the dashboard represents the same evidence used in the report.

\begin{figure*}[ht!]
\centering
\includegraphics[width=0.9\textwidth]{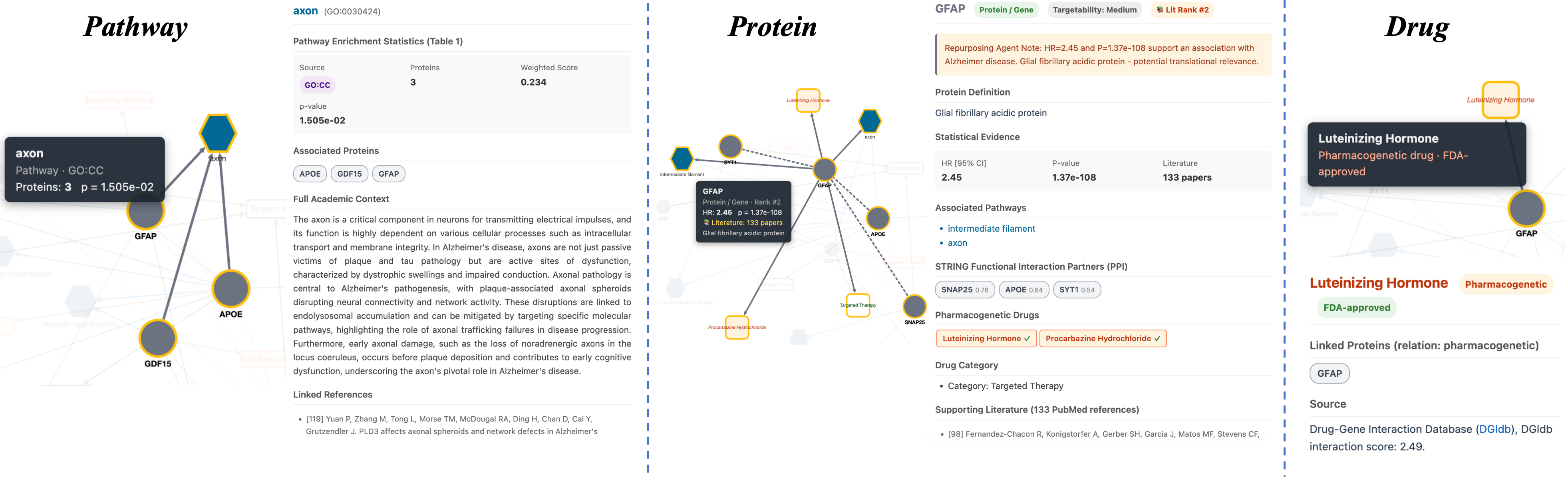}
\caption{Typed artifact flow in BioInsight. Evidence retrieval, reasoning, writing, and visualization exchange structured objects rather than only free-form text.}
\label{fig:nodes}
\end{figure*}

The reasoning-note schema used by the Reasoning Agent is shown below. Each note contains pathway-level interpretation, disease relevance, key proteins, PPI module explanations, uncertainty, and citations.
\begin{lstlisting}
{
  "pathway_id": "REAC:R-HSA-166658",
  "pathway_name": "Complement cascade",
  "disease_explanation": "...",
  "key_proteins": [
    {
      "symbol": "C3",
      "function_note": "...",
      "association": {"hr_95ci": "...", "p_value": "..."},
      "citations": ["12345678"]
    }
  ],
  "ppi_clusters": [
    {
      "proteins": ["C3", "CFH", "C4A"],
      "cluster_explanation": "...",
      "citations": ["12345678"]
    }
  ],
  "uncertainty": "...",
  "citations": ["12345678"]
}
\end{lstlisting}

This schema gives the Reasoning Agent a narrow and inspectable output channel. It must state what the pathway does, why it may matter for the disease, which input proteins drive the interpretation, which interaction modules support the mechanism, and where evidence is weak or indirect. If no relevant publications are available, citation fields remain empty and the explanation is marked as exploratory.

\subsection{Implementation Parameters}

The harness exposes model choices as configuration parameters for the Planning, Search, Reasoning, Writing, and Visualization Agents. Prompt templates are fixed across diseases; disease specificity enters through the disease name, protein table, cohort metadata, pathway terms, and retrieved evidence packets. Table~\ref{tab:key-parameters} summarizes the key non-prompt parameters used in the current configuration.

\begin{table*}[htbp]
\centering
\small
\caption{Key implementation parameters used in the current BioInsight harness.}
\label{tab:key-parameters}
\begin{tabularx}{\textwidth}{l l X}
\toprule
\textbf{Component} & \textbf{Parameter} & \textbf{Value} \\
\midrule
Pathway enrichment & Organism & \texttt{hsapiens} \\
Pathway enrichment & Significance threshold & \texttt{user\_threshold = 0.05} \\
Pathway enrichment & Correction method & \texttt{significance\_threshold\_method = "fdr"} \\
Pathway enrichment & IEA annotations & \texttt{no\_iea = True} \\
Pathway enrichment & Ordered query & \texttt{ordered = False} \\
Pathway enrichment & Highlighted terms & \texttt{highlight = True} \\
Pathway filtering & Term size range & \(10 < \texttt{term\_size} < 500\) \\
Duplicate removal & BioBERT cosine threshold & \(0.98\) \\
Pathway selection & Per-source retained pathways & \(k = 2\) \\
Report generation & Maximum analyzed pathways & 12 \\
PubMed retrieval & Maximum records per query & 50 records with abstracts \\
Semantic Scholar retrieval & Maximum records per query & 30 records \\
Publication validation & Score threshold & \(S_{\mathrm{pub}} \geq 0.25\) \\
Publication storage & Validated publications per pathway & Top 20 \\
Reasoning clusters & Retained PPI components & Top 2 connected components \\
STRING dashboard retrieval & Confidence threshold & 0.4 \\
STRING dashboard retrieval & Maximum fetched proteins & Top 80 proteins by literature and pathway evidence \\
STRING dashboard retrieval & Maximum stored edges & 60 edges per query \\
DGIdb retrieval & Queried genes & Top 8 ranked genes \\
LLM generation & Temperature & 0.7 \\
LLM retry & Default retry setting & 5 retries \\
\bottomrule
\end{tabularx}
\end{table*}

Artifacts are stored under a disease-specific result directory. The cache directory stores ranked pathway tables, selected top pathways, pathway overview drafts, iterative report drafts, citation-formatted reports, dashboard schemas, and rendered dashboard pages. External fetches are cached as JSON files so repeated runs can reuse STRING and DGIdb responses subject to each resource's licensing terms.

\subsection{Quality Control and Failure Handling}

The harness is designed to expose evidence insufficiency rather than hide it behind fluent text. Table~\ref{tab:failure-handling} summarizes common failure modes and the corresponding graceful degradation behavior.

\begin{table*}[htbp]
\centering
\small
\caption{Failure modes and graceful degradation behavior.}
\label{tab:failure-handling}
\begin{tabularx}{\textwidth}{
  >{\raggedright\arraybackslash}p{0.25\textwidth}
  >{\raggedright\arraybackslash}X
}
\toprule
\textbf{Failure mode} & \textbf{System behavior} \\
\midrule
No significant enriched pathways & Report evidence insufficiency instead of generating unsupported pathway mechanisms. \\
Enriched pathway with weak literature support & Retain the pathway in the ranked table when appropriate, but mark the interpretation as literature-weak or exploratory. \\
Missing protein function annotation & Omit the protein-level explanation or mark function evidence unavailable. \\
No protein-protein interaction edges & Skip cluster-level interpretation and analyze proteins individually. \\
STRING link unavailable & Retain the generated network image when available or omit the link if no network can be produced. \\
Open Targets or DGIdb returns no records & Do not infer therapeutic relevance from absence of records. \\
PubMed metadata lookup fails & Retain the PubMed link and surface missing metadata rather than dropping the citation. \\
Malformed Reasoning Agent JSON & Preserve the raw response for inspection, retry, or exclusion from final report assembly. \\
Writing revision damages tables, links, or image syntax & Use stored pre-revision and post-revision drafts for inspection and correction. \\
External API unavailable & Fall back to cached responses or narrower retrieval behavior when possible. \\
\bottomrule
\end{tabularx}
\end{table*}

If pathway enrichment returns no significant pathways, the system should report that the input protein set does not support a pathway-level interpretation rather than hallucinating mechanisms. If a pathway is statistically enriched but has little or no PubMed support, it can remain in the ranked table, but its narrative interpretation should be marked as literature-weak or exploratory. If UniProt-derived function annotations are missing for a protein, the protein-level explanation is omitted or marked unavailable. If no protein-protein interaction edges are found, the system avoids cluster-level interpretation and instead treats proteins individually.

External translational resources are also handled conservatively. If Open Targets or DGIdb returns no records for a protein, the system does not infer therapeutic relevance from absence of evidence. If STRING fails to return an interactive link, the report can still retain the generated network image, or omit the STRING link if no network is available. If a PubMed metadata lookup fails during citation formatting, the PubMed link is retained and missing metadata is surfaced rather than silently dropping the citation.

Language-model failures are handled at artifact boundaries. If the Reasoning Agent returns malformed JSON, the raw response is preserved as an inspectable artifact and can be retried or excluded from final report assembly. If a coherence revision damages tables, links, or image syntax, downstream parsing and manual inspection can identify the failure because the pre-revision and post-revision drafts are both stored.

\section{Experimental Setup Details}
\label{app:experimental-setup}

This appendix provides reproducibility details for the three experimental settings described in Experiment Section and the surrounding experiments.

\subsection{Baseline and Input Matching}
All systems receive the same disease names, protein association tables, and task instructions within each evaluation setting. Search-enabled baselines are given the same evidence-seeking objective and a comparable retrieval budget when applicable. BioInsight uses fixed agent prompts, pathway-ranking weights, retrieval parameters, citation-formatting rules, and dashboard-generation rules across all disease cases. Automatic metrics are computed from raw model outputs without manual correction.

For RQ1 and RQ2, we compare BioInsight with GPT-5.5, DR-Tulu-8B, Gemma-4-31B, and Qwen3.5-9B. For RQ3, we evaluate Claude Sonnet 4.6, GPT-5.5 + Search, Gemini Deep Research, Gemma-4-31B + Search, Qwen3.5-9B + Search, DR-Tulu-8B, BioInsight, and BioInsight without search decomposition. The ablated BioInsight variant keeps the report-generation stage but removes the full explicit search and reasoning decomposition, allowing us to test the contribution of the harnessed evidence workflow.

\subsection{BioASQ Phase B}
BioASQ Phase B is used as a standardized exact-answer test for yes/no, factoid, and list questions. We evaluate Batch 1 and report yes/no accuracy, yes/no macro F1, factoid strict accuracy, factoid mean reciprocal rank, and list F-measure. This setting checks whether each system can select concise biomedical entities and evidence-supported answers before those entities are expanded into longer reports and dashboards.

\subsection{BioInsight-100}
BioInsight-100 is constructed from UniProt-derived function records and STRING interaction evidence. GPT-5.5 is used to generate candidate protein-function questions that require cross-evidence reasoning, and biomedical experts select 100 challenging questions to form BioInsight-100. Each answer is scored on a 0--10 scale according to biomedical correctness, evidence use, protein-function specificity, pathway or interaction reasoning, and clarity.

An example from BioInsight-100 is shown below. The question requires the model to integrate protein-function annotations with interaction evidence and to recognize when the interaction context is weak or absent.

\begin{lstlisting}
{
  "type": "function_ppi_integration",
  "question": "How do known interactions inform CEND1's role in neuronal differentiation?",
  "answer": "Integration is limited because no high-confidence STRING associations are available; UniProt notes homodimerization, aligning with a membrane role in promoting neuronal differentiation.",
  "evidence": [
    {
      "source": "STRING",
      "confidence": "high",
      "field": "No high-confidence associations for Q8N111 in provided STRING context"
    },
    {
      "source": "UniProt",
      "confidence": "high",
      "field": "FUNCTION; SUBUNIT (homodimerization)"
    }
  ]
}
\end{lstlisting}

\subsection{Disease Cases for Report Evaluation}
\label{app:diseases}
Table~\ref{tab:diseases} summarizes the five disease cases used in end-to-end report and dashboard evaluation. The set covers neurodegenerative, psychiatric, cardiovascular, renal/metabolic, and autoimmune/inflammatory disease contexts.

\begin{table*}[htbp]
\centering
\small
\caption{Disease categories and evaluation rationale for end-to-end report and dashboard evaluation.}
\label{tab:diseases}
\resizebox{\textwidth}{!}{
\begin{tabular}{p{3.2cm} p{3.0cm} p{4.5cm} p{5.2cm}}
\hline
\textbf{Disease} &
\textbf{Type } &
\textbf{Key mechanistic features} &
\textbf{Evaluation value for BioInsight} \\
\hline
AD / Alzheimer's disease &
Neurodegenerative disease &
Protein aggregation, neuroinflammation, synaptic degeneration, aging &
Tests integration of complex neurological mechanisms and long-term disease progression evidence. \\
Depression &
Psychiatric disorder &
Neurotransmitters, inflammation, endocrine regulation, gene--environment interactions &
Tests interpretation under heterogeneous mechanisms and uncertain disease biology. \\
Atrial fibrillation and flutter &
Cardiac rhythm disorder &
Electrophysiological remodeling, cardiac structural remodeling, inflammation, coagulation risk &
Tests integration of clinical risk, biomarker, and proteomic evidence. \\
Chronic kidney disease &
Chronic metabolic/renal disease &
Inflammation, fibrosis, oxidative stress, tubular injury &
Tests interpretation of chronic progression, organ damage, and multi-pathway protein evidence. \\
Rheumatoid arthritis &
Autoimmune/inflammatory disease &
Immune-cell activation, cytokines, synovial inflammation, tissue destruction &
Tests interpretation of immune pathways and target-relevant protein evidence. \\
\hline
\end{tabular}
}
\end{table*}

\section{Human Evaluation Protocol}
\label{app:human-evaluation}

This appendix describes the human evaluation protocol used for report-level expert assessment.
The goal of the evaluation is to assess whether BioInsight produces biomedical synthesis outputs that are complete, scientifically valid, evidence-integrative, well-prioritized, and usable for biomedical researchers.

\paragraph{Evaluation setup.}
We evaluate system outputs on the five disease--protein interpretation cases in Table~\ref{tab:diseases}.
For each disease case, evaluators are shown anonymized outputs from four systems: GPT-5.5 + Search, Gemini Deep Research, DR-Tulu-8B, and BioInsight.
Each output consists of the generated biomedical interpretation report and, when available, the associated dashboard or evidence views. 
System names are hidden from evaluators, and outputs are presented in randomized order to reduce ordering and model-identity bias. 
Evaluators assign a score from 1 to 5 for each evaluation dimension and provide a brief justification for each score. We invited four domain experts for human evaluation.

\paragraph{Evaluation dimensions.}
Following our report-level expert evaluation guideline, each output is rated along five dimensions: comprehensiveness, biomedical validity, evidence grounding and traceability, prioritization and research depth, and readability/dashboard usability.

\textbf{Comprehensiveness} measures whether the system provides a complete end-to-end interpretation of the input protein list. 
This includes pathway enrichment, key proteins, molecular functions, disease relevance, molecular mechanisms, PPI or network context, therapeutic associations when relevant, citations, dashboard evidence views, and cross-links among proteins, pathways, mechanisms, and translational findings.
Reports are penalized when they cover only isolated proteins or pathways, omit important disease-protein interpretation modules, or include modules only as shallow labels.

\textbf{Biomedical validity} measures whether the biological statements, disease interpretations, pathway explanations, protein annotations, mechanism chains, drug associations, and network interpretations are factually correct, scientifically plausible, and appropriately qualified. 
Evaluators penalize hallucinated mechanisms, incorrect protein functions, generic disease associations, unsupported causal language, overextended clinical inference, fabricated or loosely related citations, and context errors such as tissue, stage, species, or biomarker-versus-causality mismatch.
Rare or non-obvious pathways are not penalized merely for being uncommon; they receive positive credit when they are biologically plausible, disease-relevant, and supported by evidence.

\textbf{Evidence grounding and traceability} measures whether conclusions are explicitly grounded in visible evidence from the report, dashboard, and intermediate artifacts. 
Evaluators check whether the output preserves an auditable chain from raw or intermediate evidence to interpretation, such as protein statistics, mapped genes, enriched pathways, pathway rankings, PPI edges, literature records, database identifiers, drug records, and citation-linked claims.
Strong outputs keep evidence close to the relevant claim, expose evidence fields such as p-values, enrichment scores, hit counts, confidence scores, literature counts, PubMed references, or drug--target records, and disclose weak, missing, indirect, or uncertain evidence.
Outputs are penalized when plausible claims are untraceable, citations are only loosely connected to claims, evidence provenance is mixed or unclear, or mechanistic conclusions exceed the strength of the cited evidence.

\textbf{Prioritization and research depth} measures whether the system identifies which proteins, pathways, mechanisms, PPI modules, biomarkers, drug links, or translational findings deserve deeper analysis using evidence-weighted biomedical reasoning. 
Evaluators consider whether rankings are supported by visible quantitative and qualitative evidence, including enrichment statistics, pathway scores, mapped genes, protein hit counts, literature support, hazard ratios, PPI confidence, database evidence, druggability, and disease specificity.
Strong outputs distinguish core data-supported findings from supporting proteins, peripheral associations, uncertain hits, indirect evidence, and speculative hypotheses.
Outputs are penalized for arbitrary ordering, significance-only ranking without biological interpretation, literature-only ranking without enrichment context, shallow lists, generic pathway discussion, or over-prioritized translational claims without traceable support.

\textbf{Readability and dashboard usability} measures whether the report and dashboard are understandable, well-structured, visually clear, and useful for biomedical evidence exploration. 
For reports, evaluators consider disease-story coherence, logical flow, terminology control, citation placement, and clarity of uncertainty. 
For dashboards, evaluators consider whether users can move from overview to detail, trace visual elements to evidence sources, drill down from pathways to proteins and citations, compare mechanisms or evidence strength, and interact with the interface without excessive cognitive load.

\paragraph{Rating scale.}
All dimensions are scored on a five-point Likert scale. 
Although each dimension has dimension-specific criteria, the general interpretation of the scale is as follows:

\begin{table}[t]
\centering
\small
\begin{tabular}{cp{6.5cm}}
\hline
\textbf{Score} & \textbf{General interpretation} \\
\hline
1 & Very poor. The output is incomplete, misleading, unsupported, or difficult to use. \\
2 & Weak. Some relevant information is present, but important components are missing, shallow, inaccurate, or poorly connected. \\
3 & Moderate. The output covers major components and is mostly understandable, but evidence integration, validity, ranking transparency, or usability remains limited. \\
4 & Strong. The output is largely complete, disease-relevant, evidence-supported, and useful, with only minor issues. \\
5 & Excellent. The output is comprehensive, scientifically reliable, deeply synthesized, well-prioritized, and easy to inspect through the report and dashboard. \\
\hline
\end{tabular}
\caption{General five-point rating scale used in expert evaluation. Dimension-specific scoring instructions are provided to evaluators.}
\label{tab:human-eval-rating-scale}
\end{table}

\paragraph{Dimension-specific scoring guidelines.}
For coverage, a score of 1 indicates that the system only lists proteins or pathways with little interpretation, while a score of 5 indicates comprehensive coverage of pathway enrichment, protein function, disease relevance, mechanisms, PPI or interaction clusters, therapeutic associations, citations, and dashboard evidence exploration. 
For biomedical validity, a score of 1 indicates severe biomedical errors or hallucinated mechanisms, while a score of 5 indicates highly accurate, disease-specific, mechanistically rigorous, and uncertainty-aware interpretation. 
For evidence synthesis depth, a score of 1 indicates no meaningful synthesis beyond lists of facts, while a score of 5 indicates a disease-specific, evidence-weighted, uncertainty-aware mechanistic model with clear therapeutic implications. 
For ranking quality, a score of 1 indicates arbitrary or misleading rankings, while a score of 5 indicates highly interpretable, evidence-weighted, disease-specific, uncertainty-aware, and actionable prioritization. 
For readability and dashboard usability, a score of 1 indicates that the report is difficult to follow or the dashboard is confusing, while a score of 5 indicates a coherent disease story and intuitive, accurate, source-traceable evidence exploration.

\paragraph{Qualitative analysis and Bias control.}
In addition to numerical scores, we analyze evaluator justifications to identify recurring strengths and failure modes. 
We group comments into categories such as incomplete evidence coverage, generic disease interpretation, unsupported causal language, weak protein-to-pathway linkage, shallow ranking explanation, citation misalignment, dashboard inconsistency, and poor evidence traceability. 
These qualitative findings are used to interpret the quantitative results above.

The evaluation is blinded with respect to system identity, but it is still limited by the number of disease cases and expert evaluators. 
Biomedical interpretation is also inherently judgment-dependent: experts may differ in how they weigh disease specificity, mechanistic plausibility, and evidence strength. 
To reduce bias, all systems are evaluated on the same disease cases, using the same scoring rubric, randomized output order, and identical evaluation forms. 
Nevertheless, the human evaluation should be interpreted as expert assessment of research utility and evidence quality, rather than as a definitive biomedical validation of every generated claim.

\section{The Use of Large Language Models (LLMs)}
In order to reduce typos during the writing process and to optimize complex sentence structures so that the article becomes simpler and easier to read, we use mainstream large language models to refine certain paragraphs. For example, we use prompts such as “Help me correct the typos and grammatical errors in the above text, and streamline the logic to make it clear and easy to understand.”

\end{document}